# Analysis of Arrhythmia Classification on ECG Dataset

Taminul Islam*
*Department of Computer Science and Engineering*
*Daffodil International University*
Ashulia, Bangladesh
taminul@ieee.org

Arindom Kundu
*Department of Computer Science and Engineering*
*Daffodil International University*
Ashulia, Bangladesh
arindom15-10557@diu.edu.bd

Tanzim Ahmed
*Department of Computer Science and Engineering*
*Daffodil International University*
Ashulia, Bangladesh
tanzim15-10801@diu.edu.bd

Nazmul Islam Khan
*Department of Computer Science and Engineering*
*Daffodil International University*
Ashulia, Bangladesh
nazmul15-13802@diu.edu.bd

*Abstract*— The heart is one of the most vital organs in the human body. It supplies blood and nutrients in other parts of the body. Therefore, maintaining a healthy heart is essential. As a heart disorder, arrhythmia is a condition in which the heart's pumping mechanism becomes aberrant. The Electrocardiogram is used to analyze the arrhythmia problem from the ECG signals because of its fewer difficulties and cheapness. The heart peaks shown in the ECG graph are used to detect heart diseases, and the R peak is used to analyze arrhythmia disease. Arrhythmia is grouped into two groups - Tachycardia and Bradycardia for detection. In this paper, we discussed many different techniques such as Deep CNNs, LSTM, SVM, NN classifier, Wavelet, TQWT, etc., that have been used for detecting arrhythmia using various datasets throughout the previous decade. This work shows the analysis of some arrhythmia classification on the ECG dataset. Here, Data preprocessing, feature extraction, classification processes were applied on most research work and achieved better performance for classifying ECG signals to detect arrhythmia. Automatic arrhythmia detection can help cardiologists make the right decisions immediately to save human life. In addition, this research presents various previous research limitations with some challenges in detecting arrhythmia that will help in future research.

## I. INTRODUCTION

The human heart sends blood to the lungs to acquire oxygen, then returns it to the blood and carries it throughout the body. The leading cause of death is cardiovascular disease (CVDs) worldwide [1]. By doing research, WHO (World Health organization) found closely 17.9 million, almost 32% of the global deaths caused by CVDs, in 2019. CVDs can be classified into three major groups – electrical (irregular heartbeats owing to malfunctions of the heart's electrical system or arrhythmia), structural (heart muscle disease or cardiomyopathy), circulatory (high blood pressure and coronary artery disease) [2]. The coronary artery blockage of the heart is responsible for the heart attack shown in Fig. 1.

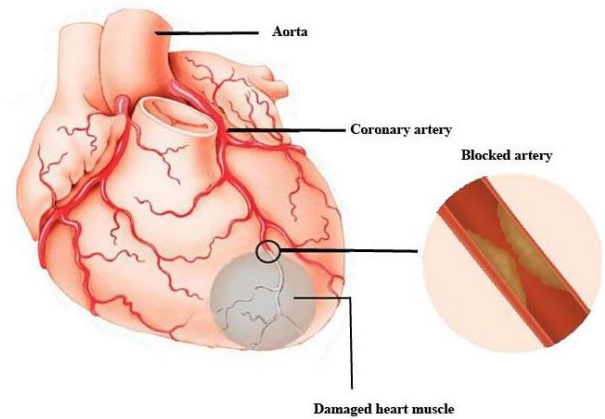

Fig. 1. Heart attack or myocardial infarction

In order to examine the myocardial electrical transmissions of the human heart in the waveform, electrocardiography (ECG) is the preferred approach. This task is done by an instrument which is called Electrocardiogram; it is known as ECG shortly. ECG is one kind of graph that shows voltage versus time of electrical movement of the heart. This graph is achieved by using electrodes placed on the skin, which detects electric signals of the heart which it beats each time [3]. ECG is generally used to detect diseases of the human heart such as a coronary artery, arrhythmias, cardiomyopathy, heart attacks. In this study, we discussed arrhythmia classification using Long Short-Term Memory (LSTM), Deep Convolutional Neural Networks (DCNNs), and Machine Learning (ML).

ECG signal usually provides some particular human heart information, such as the position of the heart and the heart chamber size. [4] It also reveals the source and spread of impulses. The ECG can visualize cardiovascular rhythm and conduction disorders. The medication effects on the heart can also be identified.

Arrhythmia is common among other heart diseases [5][6]. Irregular heartbeat is known as arrhythmia. According to the speed of the heartbeat, arrhythmia is grouped into two categories –

Tachycardia: Tachycardia is the medical term for rapid heartbeats. At least one person in the room has a heart rate of more than 100 beats per minute.

Bradycardia: Bradycardia is the medical term for irregular, slow heartbeats. There are less than 60 beats per minute in your heart.

Arrhythmia can be occurred by damage from illness, injury or genetically. Doctors suggest a common test ECG for diagnosing the rhythm of the heart [7][8]. Without treatment of the arrhythmia, enough blood may not pump by the heart to the body and it can harm the heart, brain, and other parts of the body. Therefore, cardiologists correctly identify abnormal heartbeats is essential [9].

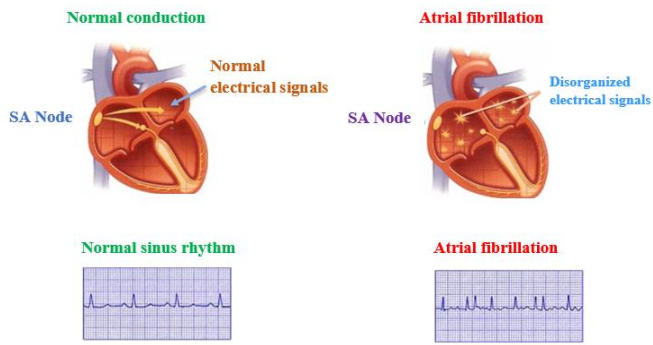

Fig. 2. Normal heart vs Irregular heart

To create an artificial intelligence system, deep learning techniques are used and artificial neural networks (ANN) is used to build this technique for doing complex analysis. Deep convolutional neural networks (DCNNs) are showing better performance on object detection [10], image classification, recommendation systems [11] and natural language processing [12]. This review study focused on arrhythmias classification on ECG based dataset using DCNN because it is becoming very popular on image classification. Besides, other techniques for detecting arrhythmia have discussed.

To uncover linked works relevant to machine learning and methodologies for deep study in the classification of arrhythmias, a good research question is important. The stages to address the correct research questions, including population, intervention, outcomes and context, are proposed by [13]. The research topic criteria are illustrated in Table 1.

TABLE 1: CRITERIA OF RESEARCH QUESTION

| Criteria | Details |
|---|---|
| Population | Medical Patient |
| Intervention | ML and DL approaches for prediction |
| Outcome | Important attributes, Accuracy and Classification |
| Context | Hospital Laboratory |

This study focused on some research questions, following the above criteria where the database can be found that have mostly been used for Arrhythmia classification work. This research shows the most common methods of using machine learning to make predictions. The present predictive models' performances and the challenges to developing a model for predicting arrhythmia classification can be found by this study.

The remaining content of this paper is structured as follows: several research methodology sections present various research works has been already done on the arrhythmia detection. In the discussion section, various challenges in arrhythmia detection are discussed and the paper concludes lastly.

## II. RELATED WORKS

For the classification of irregular heartbeats, nine layers of convolutional neural networks were suggested by Acharya et al. [14], each composed of 3 convolution layers, one fully connected layer, and one max-pooling layer. When it came to beating the hearts of the people they were studying, they classified them into five categories: Fusion(F), Non ectopic(N), V, S, and unknown beats (Q). ECG heartbeats are pulled from MIT-BIH arrhythmia database and verified by two cardiologists before being used. When the underlying dataset's Z-score is imbalanced, synthetic heartbeat data is produced by altering its mean and standard deviation, which yields an accuracy of 94% from the additional data. However, when trained on the original dataset, this model had an accuracy of 89.07 percent.

Another study [15] proposed two deep learning methods based on CNNs - end to end approach that analyzes the heartbeats directly and hierarchical process of two stage. The first step is to determine the kind of heartbeat, and the second is to use the MIT-BIH arrhythmia dataset to sort the various types of heartbeats into 15 distinct categories. Data augmentation technique GAN is used for generating synthetic heartbeat data to balance the dataset for each class. Approximately 98.30% accuracy and 90% precision are gained from the end-to-end approach. By using another approach - two stage hierarchical process, 98% accuracy and precision of 93.5% are obtained. Eleven layers of CNN were used by Abdalla et al. [16] to identify 10 distinct kinds of arrhythmias, including four convolutional layers, max pooling layers and three max pooling layers. In order to compile the statistics, we used the hospital database at "Massachusetts Institute of Technology (Beth Israel Medical Center)". After train and testing the data is amplified by using the Z-score technique, different means and standard deviation to achieve an adequate and balanced dataset. The dataset is divided into following manner- 80% for training and 20% for testing data and 99.84% accuracy is obtained.

The MIT-BIH ECG signal dataset was used by Yildirim et al. [17] to create a 16-layer one-dimensional conventional neural network (1D CNN) model for recognizing 17 rhythm types. From the dataset, one thousand (1000) signal fragments of 45 patients were selected randomly where each contains 3600 samples. For training purposes 70% (700 samples), for testing and validation purposes 15% (150 samples) data are used respectively. By applying rescaling technique, they achieved better results and achieved 91.33% accuracy for classifying 17 classes.

Eight different kinds of ECG signals from either the dataset of MIT-BIH arrhythmia was classified using a two-dimensional CNN model with the addition of Long Short-Term Memory (LSTM) by Zheng et al. [18]."Premature Ventricular Contraction (PVC)", Ventricular Flutter Waves and Ventricular Escape Beat are all signs of irregular heartbeats, as is the RBBB, LBBB and the normal sinus rhythm (NOR) (VFW). LSTM and a fully connected layer are employed to achieve a prediction accuracy of 99.01 percent in the convolutional, LSTM, and fully connected layers, respectively. Workflow diagram of is depicted in Fig. 3.

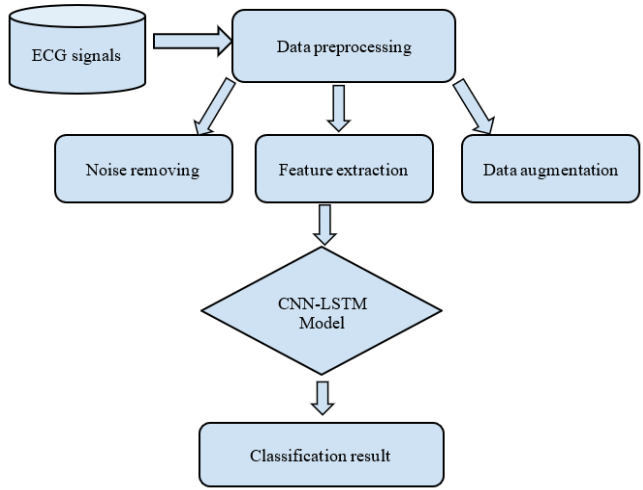

Fig. 3. Workflow of the CNN-LSTM model

Gao *et al.* [19] proposed a four layers of LSTM model containing an input layer, 2 fully connected layers and an LSTM layer. The model classifies 8 various types of beats of the MIT-BIH arrhythmia dataset. Focal loss technique is used for dealing with imbalanced data. To remove noise from the ECG signal, the Daubechies 6 (db6) discrete wavelet transform is used, then the sliding window search method is used for extracting heartbeat from the ECG signal and the Z-score technique is applied for achieving normalized data, in the preprocessing stage. In this work, the researchers trained the LSTM network using FL and an overall accuracy of 99.26% is achieved. Yildirim *et al.* [20] proposed a CAE (Convolutional Auto Encoder) with LSTM model to predict 5 several arrhythmia classes from the MIT-BIH database. The CAE model was used to compress raw ECG signal beats in order to extract coded features from each one and then these were utilized in an LSTM network to classify the arrhythmia class in the following phase. They achieved 99.23% accuracy using raw data in the LSTM model and 99.11% accuracy is obtained using coded features.

### III. ANALYTICAL EVALUATION

Normal ECG signal has five distinct peaks such as P wave, Q wave, R wave, S and T wave.

P peak (wave) - It happens during atrial depolarization. Q peak (wave) - It is a downward wave which imitates the P wave. R peak (wave) **-** It is an upward wave which imitates a P wave. S peak (wave)- It is a downward wave which imitates a R wave. T peak (wave) - It is linked with ventricular repolarization. Here PR interval refers when the depolarization waves move from the atria to the ventricles, the time interval between them is called PR interval. Besides, QT interval is in the ECG signal, this interval indicates the number of ventricular activities. Between the beginning of repolarization and depolarization of the ventricular muscle, the ST segment is known as the repolarization interval. There are three stages of depolarization in a QRS complex, which are: initiation, elongation, and termination.

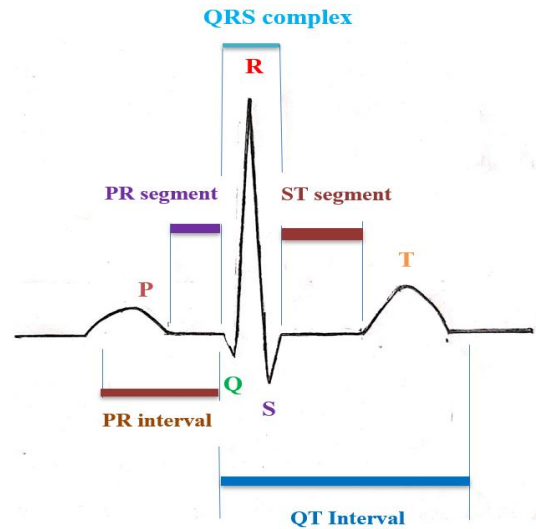

Fig. 4. The ECG signal peaks.

It is discovered in this research endeavor that there are two main datasets from which all of the relevant data has been gathered.

MIT-BIH is a database that was the maiden widely obtainable collection of standard test items for examining arrhythmias. It holds 48 records of 47 individuals with two channel ECG recordings. Arrhythmia Laboratory of the Boston's Beth Israel Hospital (BIH) collected this data during 1975 & 1979; around 60% of the individuals were inpatients [21][22].

Arrhythmia Dataset classifies cardiac rhythmic presence and absence and does this in one of the sixteen groups. There are 452 instances and a total of 279 attributes where 206 are linear valued. This is a classification work where characteristics of the attribute are integer, categorical and real. The type of this dataset is multivariate [23]. H. Altay Guveni, Burak Acar, and Haldun Muderrisoglu are the original owner of this dataset. This dataset has been applied to the work of H. Altay [28]. This data refers to Class 01 in 'normal' ECG classes. In addition to class 2 to class 15, several classes are assigned and for the remaining unclassified class 16 is assigned. This work also informs that such a classification is a computer program, but here there is a separation between the classification of cardiology's and the classification of the program. This dataset has been uploaded to UCI Machine Learning Repository Website [24].

#### A. Machine Learning Approaches

A large number of machine learning methods like SVMs and Artificial neural networks (ANNs) have been applied to detect arrhythmia [25]. Alfaras *et al.* [26] introduced a machine learning method -Echo State Networks that classifies the heartbeats of the process ECG records into 2 classes- SVEB+ and VEB+ on the basis of the morphology. Two ECG datasets are used in this work, the MIT-BIH AR that holds two leads: lead II, lead V1 and the AHA contains 2 leads: lead A and B. From this analysis three ECG records in the AHA database and four ECG records in the MIT-BIH

AR database are eliminated. In the MIT-BIH AR, the lead II provides 98.6% accuracy and 96.8% accuracy is achieved in the lead V1. In the AHA, the lead A gives 98.6% accuracy and 97.8% accuracy is obtained in the lead B. An optimized block-based NN classifier is used to identify five several categories of heartbeats and achieve an accuracy of 97% [27]. Banerjee *et al.* [28] introduced a cross wavelet transform (XWT) based method for classifying normal and abnormal ECG patterns and achieved an accuracy of 97.6%. The MIT-BIH arrhythmia database contained eight different types of ECG beats, and Jha et al. [29] used SVM classifier and TQWT-based characteristics of ECG beats to detect them with a 99.27 percent accuracy rate.

Gupta *et al.* [30] presented various techniques such as Naïve Bayes, Random Forest, SVM, Neural networks etc. for identifying 14 various arrhythmia classes. Arrhythmia dataset is obtained from the UCI Machine learning repository which contains 452 rows and 279 columns with missing values. But after eliminating the missing values, the number of columns was reduced to 252. The dataset is labeled into 16 classes by the Cardiologists. Among 16 classes, classes two to fifteen represent several kinds of arrhythmia, class 1 represents normal ECG and class 16 represents unlabeled patients. Two types of Naïve Bayes classifier are applied (binomial, multinomial) without feature reduction and achieve high train and test errors. SVM classifier with mRMR feature selection technique is used in the dataset. Bootstrapping is applied for improving the accuracy of the model. But it is unable to classify class 16. It classifies class 5 as class 1 which is wrong and an anomaly detector is used for solving this issue. They achieved 70% accuracy. Random Forest classifier with bootstrapping is used and achieved overall accuracy of 72.3%. A serial classifier composed of Random Forest and linear kernel SVM-poly degree 2 gave 77.4% accuracy. Hierarchical two Random Forest classifiers are applied on the dataset and achieved 30% error. Pattern net neural network gave 69% accuracy. The summary of the results is shown in the Fig. 5.

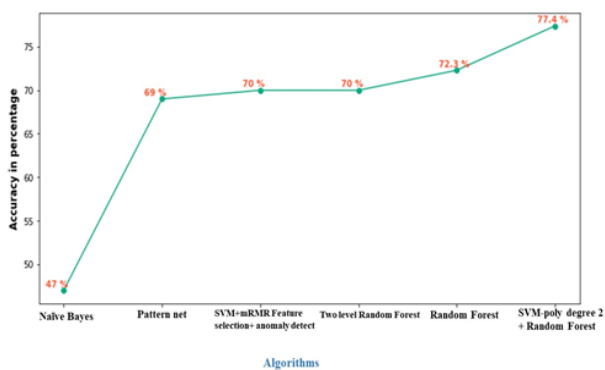

Fig. 5. Summary of the results

Devdas et al. [31] used 3 ML algorithms: SVM, Naïve Bayes and KNN to predict 16 several types of arrhythmias. Dataset is obtained from the UCI machine learning repository which contains 452 records with 280 attributes. They applied a condition (if the number of missing values is greater than 20, then select missing feature and remove it) for removing the missing value and 2 features were removed. They used the Mice method for data imputation: predicted values will fill in the gaps and Boruta method for removing unnecessary features. Dataset is divided into the following manner: 80% for training with ten cross validations. They achieved 61.5%, 46.1 % and 59.3% accuracy from SVM, Naïve Bayes and KNN algorithms before using feature selection techniques. But after applying feature selection technique, the accuracies of SVM, Naïve Bayes and KNN algorithm were 71.4%, 70.3% and 62.6%. The researcher showed that feature selection technique is essential for enhancing the result of the model.

### B. Comparative analysis

Different machine and deep learning techniques are discussed in the part of Related work and Machine Learning approaches for classification of the arrhythmia problem - Deep CNN, LSTM, CNN with LSTM, cross wavelet transform, block based NN classifier, SVM classifier, etc. Among them, hybrid models provide better performance than single models. Various techniques are used in several research works for arrhythmia classification, shown in Table 2.

TABLE 2: COMPARISON BETWEEN VARIOUS METHODS FOR CLASSIFYING SEVERAL KINDS OF ARRHYTHMIA

| No. | Comparison | | | |
|---|---|---|---|---|
| | Authors | Classes | Classifier | Accuracy |
| 01 | Acharya et al. | 5 classes | Nine layers Deep CNN | Without noise: 94.03% <br><br> With noise: 89.07% |
| 02 | Alfaras et al. | 2 classes | Echo State Network | MIT-BIH AR- <br><br> Lead II: 98.6% <br><br> Lead V1: 96.8% <br><br> AHA - <br><br> Lead A: 98.6% <br><br> Lead B: 97.8% |
| 03 | Abdalla et al. | 10 classes | Eleven layers CNN | 99.84% |
| 04 | Zheng et al. | 8 classes | CNN + LSTM | 99.01% |
| 05 | Gao et al. | 8 classes | LSTM +FL | 99.26% |
| 06 | Shaker et al. | 15 classes | Two DCNNs approaches: End-to-end Two stages hierarchical | End-to-end approach: 98.30% Two-stage process: 98% |
| 07 | Jha et al. | 8 classes | SVM +TQWT | 99.27% |
| 08 | Banerjee et al | Normal and abnormal ECG pattern | XWT | 97.6%. |
| 09 | Sahoo et al | 5 classes | NN classifier | 97% |
| 10 | Yildirim et al. | 5 classes | CAE + LSTM | Using raw data: 99.23% Using Coded features: 99.11% |

### IV. DISCUSSION

With the further evolution of machine and deep learning, MIT-BIH dataset has been used on ECG arrhythmia classification. From the literature review, it is found that most of the researchers were working on the MIT-BIH dataset, which contains 48 annotated ECG records and is imbalanced. Researchers applied various techniques to balance the dataset because it will impact on the accuracy of the model. The researcher Shaker used the GAN method to create synthetic data in order to balance the dataset and achieved better results, 0.5% accuracy and 8.64% precision was increased.

On the basis of table 2, that represents ten (10) better model collections on arrhythmia detection, the CNN model achieved the highest accuracy rather than other techniques of machine and deep learning. Most of the researchers follow some common steps for detecting arrhythmia. Collecting the data of ECG is the main considering part along with preprocessing of the dataset. Researchers always focuses on Balanced dataset. Without Balanced dataset feature extraction and train of the model can't be done smoothly. In arrhythmia detection, there are several difficulties. The researchers proposed various methods to overcome those problems. ECG signals are different for individual patients because of ECG signals morphological criteria and temporal behavior. To overcome these kinds of difficulties, the researcher Shadmand applied the BBNN classifier because of its dynamic problem handling criteria. Acharya proposed a CNN model which needs long training time and is computationally costly. Another researcher proposed the LSTM model with the FL which needs a lot of time in training steps. In Deep learning approaches, interpretation is difficult and needs a big dataset to acquire better results. But it selects important features automatically. The automatic system of detecting disease will help the doctors to make the right decision.

### V. CONCLUSIONS

Heart is the engine of life. During each heartbeat, the ECG detects any changes in the heart's electrical activity and measures the heart's pace and regularity. It is necessary that the output of the Electrocardiogram should be high quality as well as accurate in order to heart related issues should be accurately identified. In this review work, we have discussed the arrhythmia, MIT-BIH ECG signals database, and research works on arrhythmia detection. Researchers applied various attempts for detecting arrhythmia problems using the MIT-BIH ECG dataset, worked with various techniques and achieved promising results. This research work also discussed the difficulties for detecting Abnormal heart beats. Researchers worked with those challenges and proposed various techniques to sort out those problems, but need more work to find better outcomes on detecting heart diseases. Abnormal heartbeats detection has to be perfect because it is connected with human life, more research work has to be done for building the almost proper system to detect the disease in the short time.

REFERENCES

[1] Cardiovascular diseases (CVDs). Address: "https://www.who.int/news-room/fact-sheets/detail/cardiovascular-diseases-(cvds)". Accessed on: 13 January, 2021.
[2] Heart Rhythm Disorder. Address: "https://upbeat.org/heart-rhythm-disorders". Accessed on: 13 January, 2021.
[3] Shaker, A.M., Tantawi, M., Shedeed, H.A. and Tolba, M.F., 2020. Generalization of convolutional neural networks for ECG classification using generative adversarial networks. IEEE Access, 8, pp.35592-35605.
[4] Sanamdikar, S. T., Hamde, S. T., & Asutkar, V. G. (2015). A literature review on arrhythmia analysis of ECG signal. International Research Journal of Engineering and Technology, 2(3), 307-312.
[5] Zheng, Z., Chen, Z., Hu, F., Zhu, J., Tang, Q. and Liang, Y., 2020. An automatic diagnosis of arrhythmias using a combination of CNN and LSTM technology. Electronics, 9(1), p.121.
[6] Abdou, A.D., Ngom, N.F. and Niang, O., 2020. Arrhythmias Prediction Using an Hybrid Model Based on Convolutional Neural Network and Nonlinear Regression. International Journal of Computational Intelligence and Applications, 19(03), p.2050024.
[7] Acharya, U.R., Oh, S.L., Hagiwara, Y., Tan, J.H., Adam, M., Gertych, A. and San Tan, R., 2017. A deep convolutional neural network model to classify heartbeats. Computers in biology and medicine, 89, pp.389-396.

[8] Zheng, Z., Chen, Z., Hu, F., Zhu, J., Tang, Q. and Liang, Y., 2020. An automatic diagnosis of arrhythmias using a combination of CNN and LSTM technology. Electronics, 9(1), p.121.
[9] Acharya, U.R., Oh, S.L., Hagiwara, Y., Tan, J.H., Adam, M., Gertych, A. and San Tan, R., 2017. A deep convolutional neural network model to classify heartbeats. Computers in biology and medicine, 89, pp.389-396.
[10] Dhillon, A., Verma, G.K. Convolutional neural network: a review of models, methodologies and applications to object detection. Prog Artif Intell 9, 85–112 (2020).
[11] Da'u, A. and Salim, N., 2020. Recommendation system based on deep learning methods: a systematic review and new directions. Artificial Intelligence Review, 53(4), pp.2709-2748.
[12] Giménez, M., Palanca, J. and Botti, V., 2020. Semantic-based padding in convolutional neural networks for improving the performance in natural language processing. A case of study in sentiment analysis. Neurocomputing, 378, pp.315-323.
[13] B. Kitchenham et al., "Systematic literature reviews in software engineering–a tertiary study," Information and software technology, vol. 52, no. 8, pp. 792-805, 2010.
[14] Acharya, U.R., Oh, S.L., Hagiwara, Y., Tan, J.H., Adam, M., Gertych, A. and San Tan, R., 2017. A deep convolutional neural network model to classify heartbeats. Computers in biology and medicine, 89, pp.389-396.
[15] Shaker, A.M., Tantawi, M., Shedeed, H.A. and Tolba, M.F., 2020. Generalization of convolutional neural networks for ECG classification using generative adversarial networks. IEEE Access, 8, pp.35592-35605.
[16] Abdalla, F.Y., Wu, L., Ullah, H., Ren, G., Noor, A., Mkindu, H. and Zhao, Y., 2020. Deep convolutional neural network application to classify the ECG arrhythmia. Signal, Image and Video Processing, 14(7), pp.1431-1439.
[17] Yıldırım, Ö., Pławiak, P., Tan, R.S. and Acharya, U.R., 2018. Arrhythmia detection using deep convolutional neural network with long duration ECG signals. Computers in biology and medicine, 102, pp.411-420.
[18] Zheng, Z., Chen, Z., Hu, F., Zhu, J., Tang, Q. and Liang, Y., 2020. An automatic diagnosis of arrhythmias using a combination of CNN and LSTM technology. Electronics, 9(1), p.121.
[19] Gao, J., Zhang, H., Lu, P. and Wang, Z., 2019. An effective LSTM recurrent network to detect arrhythmia on imbalanced ECG dataset. Journal of healthcare engineering, 2019.
[20] Yildirim, O., Baloglu, U.B., Tan, R.S., Ciaccio, E.J. and Acharya, U.R., 2019. A new approach for arrhythmia classification using deep coded features and LSTM networks. Computer methods and programs in biomedicine, 176, pp.121-133.
[21] MIT-BIH Arrhythmia database (Online). Address: "https://archive.physionet.org/physiobank/database/mitdb/". Accessed on: 13 January, 2021.
[22] Moody, G.B. and Mark, R.G., 2001. The impact of the MIT-BIH arrhythmia database. IEEE Engineering in Medicine and Biology Magazine, 20(3), pp.45-50.
[23] UCI machine learning repository arrhythmia dataset. Address: "https://archive.ics.uci.edu/ml/datasets/Arrhythmia" (online) Accessed on: 13 January, 2021.
[24] UCI Machine Learning Repository. Address: "https://archive.ics.uci.edu/ml/index.ph". Accessed on: 31 December, 2021.
[25] Sahoo, S., Dash, M., Behera, S. and Sabut, S., 2020. Machine learning approach to detect cardiac arrhythmias in ECG signals: A survey. IRBM, 41(4), pp.185-194.
[26] Alfaras, M., Soriano, M.C. and Ortín, S., 2019. A fast machine learning model for ECG-based heartbeat classification and arrhythmia detection. Frontiers in Physics, 7, p.103.
[27] Shadmand, S. and Mashoufi, B., 2016. A new personalized ECG signal classification algorithm using block-based neural network and particle swarm optimization. Biomedical Signal Processing and Control, 25, pp.12-23.
[28] Banerjee, S. and Mitra, M., 2013. Application of cross wavelet transform for ECG pattern analysis and classification. IEEE transactions on instrumentation and measurement, 63(2), pp.326-333.
[29] Jha, C.K. and Kolekar, M.H., 2020. Cardiac arrhythmia classification using tunable Q-wavelet transform based features and support vector machine classifier. Biomedical Signal Processing and Control, 59, p.101875.
[30] Gupta, V., Srinivasan, S. and Kudli, S.S., 2014. Prediction and classification of cardiac arrhythmia.
[31] Devadas, R. M (2021). Cardiac arrhythmia classification using svm, knn and naive bayes algorithms. International Research Journal of Engineering and Technology (IRJET),08(5),3937–3941. https://www.irjet.net/archives/V8/i5/IRJET-V8I5721.pdf
[32] Moody, G.B. and Mark, R.G., 2001. The impact of the MIT-BIH arrhythmia database. IEEE Engineering in Medicine and Biology Magazine, 20(3), pp.45-50.
[33] H. Altay Guvenir, Burak Acar, Gulsen Demiroz, Ayhan Cekin "A Supervised Machine Learning Algorithm for Arrhythmia Analysis." Proceedings of the Computers in Cardiology Conference, Lund, Sweden, 1997.
[34] L. Goldberger and E. Goldberger, Clinical Electrocardiography A Simplified Approach, 5th ed. StLouis, MO: Mosby, 1994, vol. I, p.341.
[35] MIT-BIH Database distribution, Massachusetts Institute of Technology, 77 Massachusetts Avenue, Cambridge, MA02139, 1998.http://www.physionet.org/physiobank/database/mitdb/. Accessed on: 13 January, 2021
[36] Isin, A. and Ozdalili, S., 2017. Cardiac arrhythmia detection using deep learning. Procedia computer science, 120, pp.268-275.
[37] E.J.S. Luza, W.R. Schwartzb, G. C. Cháveza, D. Menottia, "ECG-based heartbeat classification for arrhythmia detection: A survey", Computer Methods and Programs in Biomedicine127, pp.144-164, 2016.
[38] Vishakha S and Naik Dessai. "Review on arrhythmia detection using signal processing", International Conference on Recent Trends in Engineering, Science and Management, pp. 1573-1579. 2017
[39] Parvaneh, S., Rubin, J., Babaeizadeh, S. and Xu-Wilson, M., 2019. Cardiac arrhythmia detection using deep learning: A review. Journal of electrocardiology, 57, pp.S70-S74.
[40] Miron Kursa and Witold Rudnicki.: Feature Selection with the Boruta Package. Journal of Statistical Software, Articles, vol.36, no.11, year (2010)
[41] Rawat, W. and Wang, Z., 2017. Deep convolutional neural networks for image classification: A comprehensive review. Neural computation, 29(9), pp.2352-2449.
[42] Maggiori, E., Tarabalka, Y., Charpiat, G. and Alliez, P., 2016. Convolutional neural networks for large-scale remote-sensing image classification. IEEE Transactions on Geoscience and Remote Sensing, 55(2), pp.645-657.
[43] Jmour, N., Zayen, S. and Abdelkrim, A., 2018, March. Convolutional neural networks for image classification. In 2018 International Conference on Advanced Systems and Electric Technologies (IC_ASET) (pp. 397-402). IEEE.
[44] Sharif, O., Islam, M. R., Hasan, M. Z., Kabir, M. A., Hasan, M. E., AlQahtani, S. A., & Xu, G. (2021). Analyzing the Impact of Demographic Variables on Spreading and Forecasting COVID-19. Journal of Healthcare Informatics Research, 1-19.
[45] Islam, T., Kundu, A., Islam Khan, N., Chandra Bonik, C., Akter, F., & Jihadul Islam, M. (2022). Machine Learning Approaches to Predict Breast Cancer: Bangladesh Perspective. In International Conference on Ubiquitous Computing and Intelligent Information Systems (pp. 291-305). Springer, Singapore.